# WaveForward: An Omnidirectional Passive Wheeled Quadruped Robot with Casters

Chuanlin Zhao, Qifeng Zheng, Shuhan Wang, Tiancheng Ma, Weixian Lin, Xin Luo*, *IEEE*

*Abstract*— **Wheeled-legged robots possess both agile mobility for traversing complex terrains and high efficiency, making them suitable for long-distance transportation applications. Conventional actuated wheeled robots require specialized hardware and electrical design due to the incorporation of wheel components. We propose a novel and low-cost passive wheeled legged robot equipped with standard casters on each leg to obtain omnidirectional mobility. The control method employs an asymmetric actor-critic structure, enabling the utilization of the privileged information of the passive caster's angles and velocities. We develop a caster base posture adjustment strategy based on velocity commands, utilizing actuated joints to modify the caster base joint axis posture and thereby adjust the propulsion direction of the casters. Moreover, we implemented multiple propulsion modes to achieve varying degrees of caster twisting oscillation, converting these into propulsive force. We conducted a slalom test and mode switch experience, which shows the passive wheeled quadruped could achieve omnidirectional movement versatility, and reduce the cost of transport (COT) by up to 89.1% with respect to legged motion.**

## I. Introduction

Compared to legged robots [1], [2], the wheeled robots have shown more suitable performance in terms of transportation efficiency on smooth, flat terrain. Wheeled-legged robots with actuated wheels [3], [4] are a tradeoff between high efficiency and terrain adaptability. However, the actuated wheels assembled to the distal ends of the legs increase the weight and require major hardware modifications. Passive wheeled-legged robots with passive wheels are a supplementary scheme for long-distance flat and gentle slope terrain transportation tasks. Unfortunately, existing passive wheeled-legged robots, which incorporate only a fixed wheel with single degree of freedom (DOF) on each leg, lack omnidirectional movement capability due to the wheels' non-holonomic constraints.[5], [6], [7].

This paper introduces a novel passive wheeled-legged robot, with each leg equipped with two DOF casters, enabling omnidirectional mobility, high efficiency, and high-speed motion capabilities. The incorporation of casters renders the system strongly underactuated. To address this challenge, we propose a posture adjustment strategy for the caster base, which adjusts its posture based on velocity reference commands. This approach enables the caster to generate a constraint-based propulsive force and alleviates the severe underactuation problem. Furthermore, as each caster's oscillatory motion can generate propulsion, we designed multiple drive modes to utilize a specified number of casters for propulsion. We introduce an RL-based motion controller by leveraging an asymmetric actor-critic structure, enabling the utilization of the privileged information of the passive caster's angles and velocities. Our main contributions are as follows:

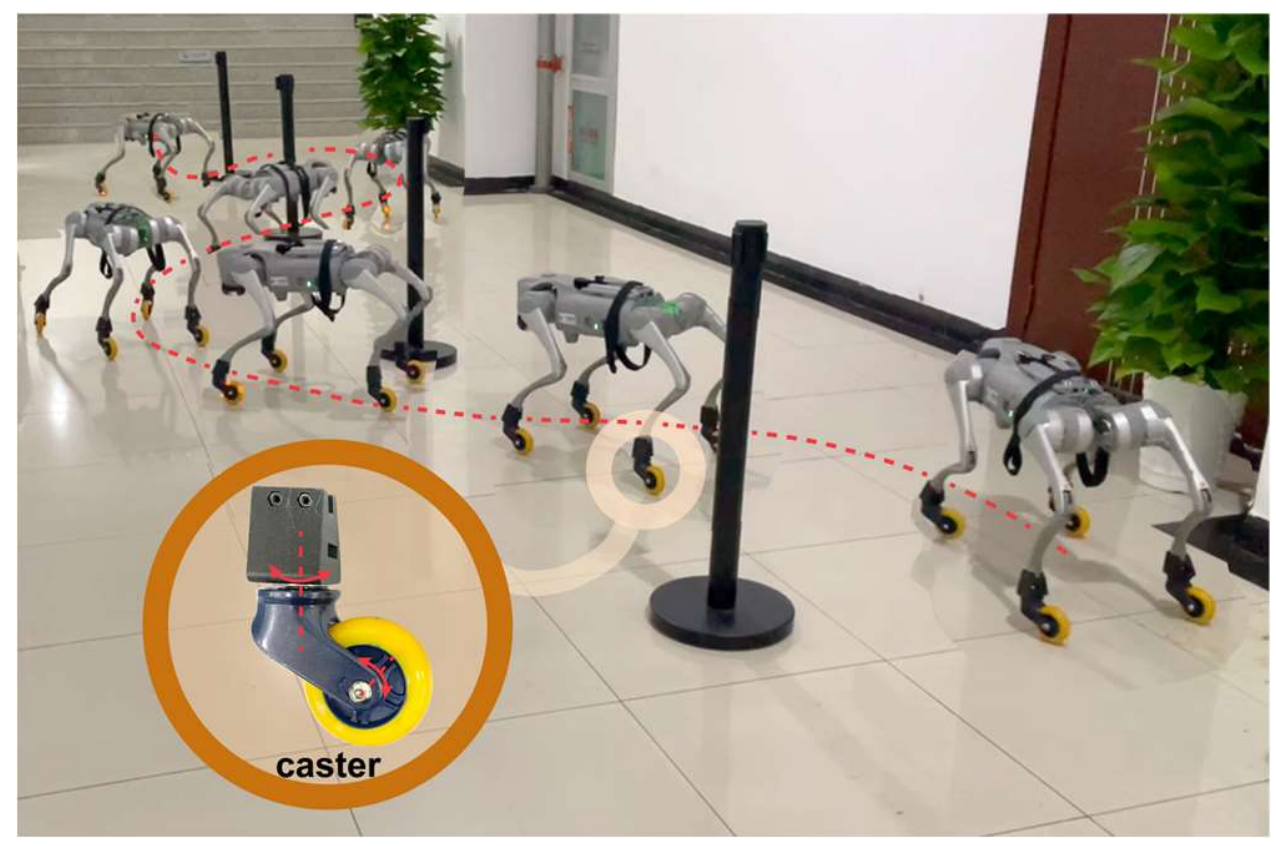

Fig. 1. The snapshot of the slalom test.

1) We develop a caster base posture adjustment strategy through actuated joints to address the propulsion force generation issue in the strongly underactuated system caused by the passive casters.
2) We introduce multiple modes to achieve varying numbers of leg twisting oscillations for conversion into propulsive forces, while enabling smooth transitions between modes.
3) We propose an omnidirectional wheeled quadruped robot based on 2-DOF casters. We verify our controller through omnidirectional movement, high-speed motion, and COT experiments.

## II. Literature Review

Wheeled-legged robots can be divided into two categories according to the wheel-driven configuration: actuated and passive wheeled-legged robots. For the actuated

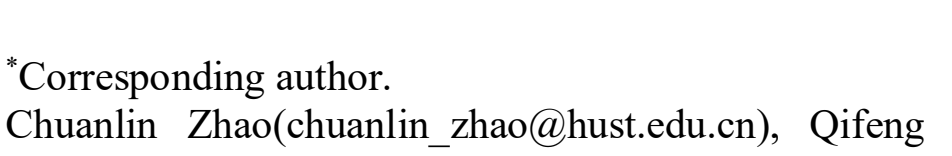

*Corresponding author.

Chuanlin Zhao(chuanlin_zhao@hust.edu.cn), Qifeng Zheng, Shuhan Wang, Tiancheng ma, Weixian Lin, Xin Luo. (mexinluo@hust.edu.cn). The authors are with the State Key Laboratory of Intelligent Manufacturing Equipment and Technology, School of Mechanical Science and Engineering, Huazhong University of Science and Technology, Wuhan 430074, China wheeled-legged robots, the integration of actuated wheels enhances the locomotion efficiency on flat terrain while maintaining the rough terrain traversal capability. The biped-wheeled robot Handle from Boston Dynamics could go through slopes, stairs, and jump over high platforms [8]. The wheeled quadruped robot ANYmal from ETH Zurich achieves adaptive locomotion control over various terrains and translates between walking and driving modes smoothly [9]. The DRC-HUBO+ combines actuated wheels on the shank and passive wheels on the foot, obtaining wheeled and legged locomotion capacities in separated control modes [10]. Similarly, the BHR-WI, attaching actuated wheels to the distal ends of legs, moves quickly and performs excellent adaptation in unmodeled rough terrains [11]. Moreover, Zhou et al. [12] designed a wheeled-legged quadruped robot, Max, which assembled actuated wheels on the knees, enabling the robot to preserve the motion agility as a quadruped while gaining the energy efficiency of wheeled robots.

Different from actuated wheels, the passive wheels normally do not need complex mechanical and electronic modifications [13]. The passive wheeled-legged robots are a feasible technical route to decrease energy consumption for long-distance transportation tasks. Chen et al. [14] achieve a humanoid roll-skating mode of a quadruped utilizing passive wheels installed on the feet. Due to the installation of a single DOF wheel, lateral linear motion cannot be achieved. Yang et al. [15] designed a novel six-legged robot for skating on ice, flat and compact terrains, equipped with two wheeled-based skateboards on four feet. The skateboards are always in the stance phase, which is called non-holonomic constraints, resulting in difficulty moving along lateral movement. Bjelonic et al. [16] realize skating motions with a force-controlled quadrupedal robot, which coordinates the passive wheeled motion and legged motion. Chen et al. [17] design a quadrupedal robot with each leg having four actuated joints and one passive wheel to enable various roller-skating locomotion. The motion-driven force is generated by the oscillatory rotation of the joint motor located in the calf. Takasugi et al. propose a three-dimensional walking and skating motion generation method to achieve sequential walking and skating motion with a skateboard and roller skate on the humanoid JAXON [21].

## III. Mechanical Design and Driving Mechanism

The omnidirectional passive wheeled quadruped robot consists of a floating base, twelve actuated revolute joints, and eight passive revolute joints. As shown in Fig. 1, each leg contains three actuated joints—hip abduction/adduction, hip flexion/extension, and knee flexion/extension—and is equipped with a replaceable caster. Each caster has two rotational axes: the swivel axis, about which the caster fork rotates relative to the caster base, and the wheel axle, also referred to as the wheel rotation axis, about which the wheel rotates relative to the caster fork. The three actuated DOF of each leg allow the pose of the caster base to be adjusted through joint actuation. Unlike a waveboard, whose caster base orientation is fixed relative to the board and therefore restricts its motion mainly to the forward direction, the proposed robot can reconfigure its caster base poses to achieve more flexible omnidirectional mobility, including lateral and backward motion.

Since the omnidirectional mobility of the robot originates from the reconfiguration of individual caster modules, we first analyze the propulsion mechanism of a single passive caster with an inclined swivel axis. A body-fixed coordinate frame $(\boldsymbol{i}, \boldsymbol{j}, \boldsymbol{k})$ is established, where $\boldsymbol{i}$ denotes the unit vector along the forward direction of the body, $\boldsymbol{j}$ denotes the unit vector along the lateral direction, and $\boldsymbol{k}$ denotes the vertically upward unit vector. The ground plane is spanned by $\boldsymbol{i}$ and $\boldsymbol{j}$, with $\boldsymbol{k}$ as its normal direction. Let $\alpha$ be the inclination angle of the caster swivel axis with respect to the vertical direction $\boldsymbol{k}$, and let $\phi$ be the deflection angle of the caster about the swivel axis.

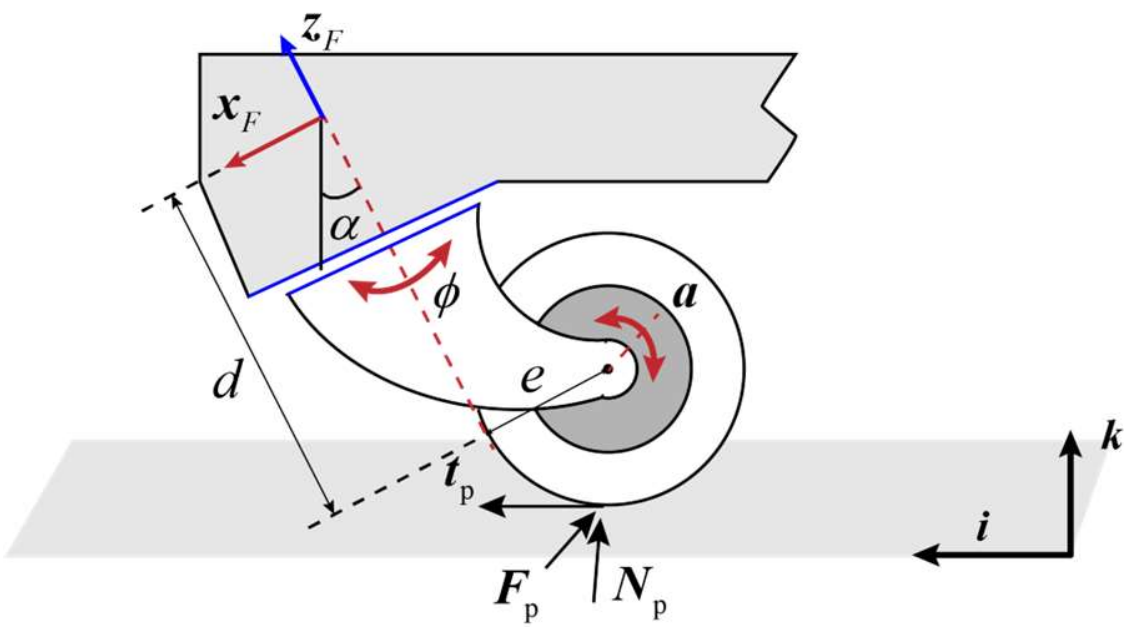

Fig. 2. Force analysis of the caster during propulsion.

Assuming pure rolling and no lateral slip at the wheel-ground contact point on a horizontal ground plane, the wheel axle direction defines the lateral no-slip direction, while the admissible rolling direction is perpendicular to the axle direction within the contact plane. Let $\boldsymbol{a}$ denote the unit vector along the wheel axle. In the body-fixed coordinate frame, it can be expressed as

$$\boldsymbol{a} = \cos\alpha \sin\phi \ \boldsymbol{i} + \cos\phi \ \boldsymbol{j} - \sin\alpha \sin\phi \ \boldsymbol{k}$$

Since the wheel axle is generally not parallel to the ground plane, its projection onto the ground plane is used to determine the non-rolling direction at the contact point. Let $\boldsymbol{a}_\mathrm{p}$ denote the projection of $\boldsymbol{a}$ onto the ground plane. Then,

$$\boldsymbol{a}_\mathrm{p} = \cos\alpha \sin\phi \ \boldsymbol{i} + \cos\phi \, \boldsymbol{j}$$

By normalizing $\boldsymbol{a}_\mathrm{p}$ , the unit vector $\boldsymbol{n}_\mathrm{p}$ along the non-rolling direction at the contact point can be obtained as

$$\boldsymbol{n}_\mathrm{p} = \frac{\cos\alpha \sin\phi \ \boldsymbol{i} + \cos\phi \, \boldsymbol{j}}{\sqrt{\cos^2\alpha \sin^2\phi + \cos^2\phi}}$$

Here, $\boldsymbol{n}_\mathrm{p}$ represents the direction in which lateral slip is constrained, and therefore it is also the direction of the no-slip constraint reaction exerted by the ground on the caster. Correspondingly, the unit vector $\boldsymbol{t}_\mathrm{p}$ along the allowable rolling direction lies in the ground plane and is perpendicular to $\boldsymbol{n}_\mathrm{p}$, which can be written as

$$t_{\mathrm{p}} = \frac{\cos\phi\,\boldsymbol{i} - \cos\alpha\sin\phi\boldsymbol{j}}{\sqrt{\cos^2\alpha\sin^2\phi + \cos^2\phi}}$$

Thus,

$$\boldsymbol{n}_{\mathrm{p}} \cdot \boldsymbol{t}_{\mathrm{p}} = 0$$

Let $\boldsymbol{F}_{\mathrm{G}}$ be the resultant ground reaction force acting on the caster. It can be decomposed into the vertical normal support force and the no-slip constraint reaction as

$$\boldsymbol{F}_{\mathrm{G}} = \boldsymbol{F}_{\mathrm{p}} + \boldsymbol{N}_{\mathrm{p}}$$

where $\boldsymbol{N}_{\mathrm{p}}$ is the magnitude of the vertical normal force exerted by the ground, and $\boldsymbol{F}_{\mathrm{p}}$ is the magnitude of the no-slip constraint reaction along $\boldsymbol{n}_{\mathrm{p}}$. Let $\boldsymbol{P}_{\mathrm{p}}$ denote the position vector from the base point of the swivel axis to the wheel-ground contact point $\mathrm{p}$, and let $\boldsymbol{z}_{\mathrm{F}}$ denote the unit vector along the swivel axis. By neglecting the friction and rotational inertia about the swivel axis, the moment component of the contact force about the swivel axis should vanish, namely

$$(\boldsymbol{P}_{\mathrm{p}} \times \boldsymbol{F}_{\mathrm{G}}) \cdot \boldsymbol{z}_{\mathrm{F}} = 0$$

Substituting the caster geometry and the force decomposition into the above moment equilibrium equation gives

$$F_{\mathrm{p}} = \frac{N_{\mathrm{P}}\sin\alpha\sin\phi}{\sqrt{\cos^2\alpha\sin^2\phi + \cos^2\phi}}$$

This result indicates that the vertical normal load $N_{\mathrm{P}}$ can induce a constraint reaction along the non-rolling direction through the combined effect of the swivel-axis inclination alpha and the caster deflection angle $\phi$. The forward component of this constraint reaction contributes to the propulsion of the body. Let $F$ denote the propulsive force component along the forward direction $\boldsymbol{i}$. Then,

$$F = \boldsymbol{F}_{\mathrm{p}} \cdot \boldsymbol{i} = F_{\mathrm{p}}(\boldsymbol{n}_{\mathrm{p}} \cdot \boldsymbol{i})$$

Substituting $\boldsymbol{F}_{\mathrm{p}}$ and $\boldsymbol{n}_{\mathrm{p}}$ into the above equation yields

$$F = \frac{N_{\mathrm{p}}\sin\alpha\cos\alpha\sin^2\phi}{\cos^2\alpha\sin^2\phi + \cos^2\phi}$$

where $F$ denote the forward propulsive force generated by a single caster. The expression indicates that the propulsion effect of an inclined-axis caster is not produced by active actuation, but by the geometric-constraint coupling among the swivel-axis inclination, caster deflection, and the no-slip wheel-ground contact constraint. Specifically, the inclined swivel axis and the caster deflection convert the vertical normal load into a static-friction-induced constraint reaction, whose forward component contributes to propulsion.

To further illustrate the influence of the two angular parameters, the relationship between the propulsive force $F$, the inclination angle $\alpha$, and the deflection angle $\phi$ is shown in Fig. 3, with the normal support force set to $N_{\mathrm{p}} = 50N$. Within the considered inclination range, the sign of $F$ is mainly determined by the sign of $\alpha$: when $\alpha > 0$, $F > 0$, whereas when $\alpha < 0$, $F < 0$. When $\alpha = 0$, the swivel axis is perpendicular to the ground, and the ground static friction only provides grip to satisfy the no-slip constraint without producing a propulsive component. Thus, no propulsive force can be generated. Moreover, $F = 0$ when $\phi = 0$, even if $\alpha \neq 0$, indicating that caster deflection is also necessary to activate the geometric-constraint coupling mechanism.

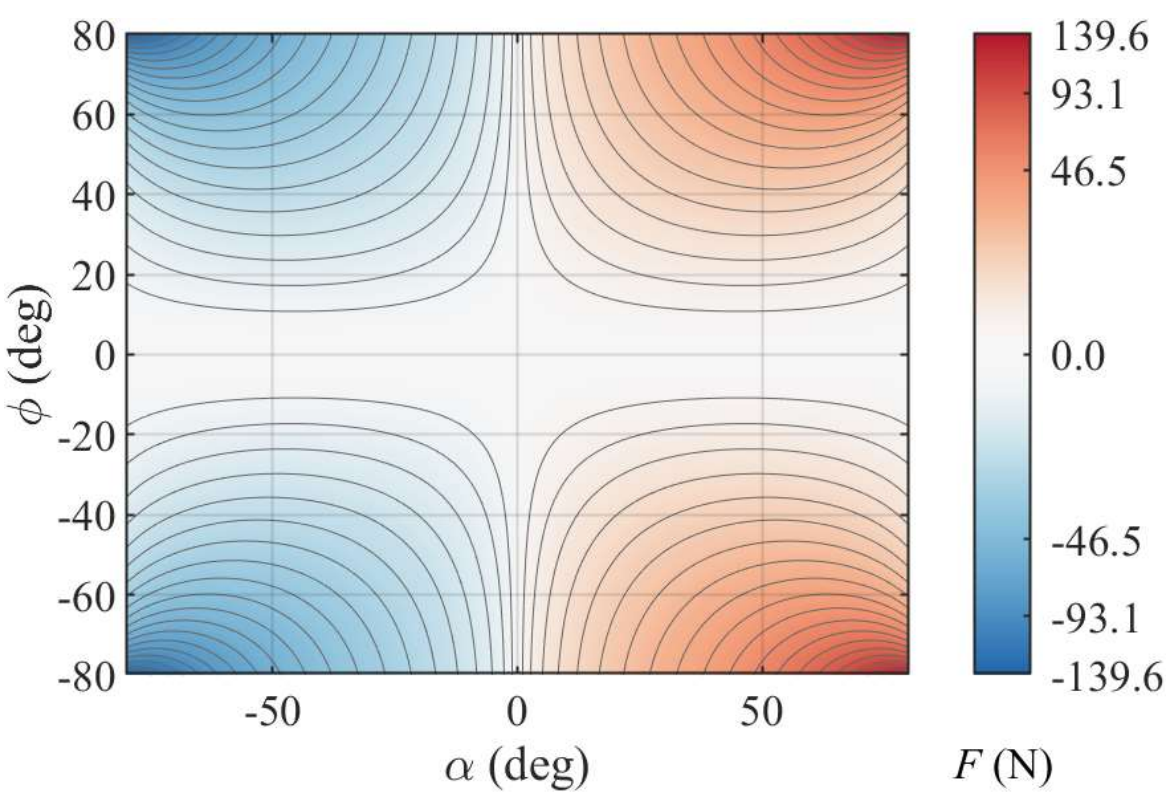

Fig. 3. Variation of propulsive force $F$ with swivel-axis inclination $\alpha$ and caster deflection $\phi$.

For a passive caster, however, the deflection angle $\phi$ tends to decrease as the wheel approaches a pure-rolling equilibrium under the wheel-ground contact constraint. Therefore, in practical locomotion, periodic variation of $\phi$ induced by leg motion or body oscillation is needed to continuously activate the propulsion mechanism. For a given $\alpha > 0$, both $\phi > 0$ and $\phi < 0$ can generate a positive forward propulsive component, while their lateral constraint forces act in opposite directions. Thus, alternating deflection of $\phi$ helps maintain an average forward propulsion while reducing lateral force effects. By adjusting the posture of the caster base, the sign and direction of the propulsive force can be controlled [19],[20].

## IV. METHODS

### A. Control Framework

The omnidirectional passive wheeled quadruped control framework is shown in Fig. 4. During the training phase, the Actor and Critic are trained using Proximal Policy Optimization (PPO) [18] within the Asymmetric Actor-Critic framework.

In the absence of encoders on the casters in real-world applications, the caster deflection angle $\boldsymbol{\phi} \in \mathbb{R}^4$ and their velocities $\dot{\boldsymbol{\phi}} \in \mathbb{R}^4$ are crucial for caster propulsion in the training phase. Consequently, we incorporate them as privileged observation inputs to the Critic. The Actor's input in the deployment stage includes mode, reference velocity commands, IMU and encoder of actuated joints. From the encoder, the joint angles $\boldsymbol{\theta}_a \in \mathbb{R}^{12}$, the joint velocities $\dot{\boldsymbol{\theta}}_a \in \mathbb{R}^{12}$ are obtained. The angular velocities $\boldsymbol{\omega} \in \mathbb{R}^3$ and the quaternion $\boldsymbol{q} \in \mathbb{R}^4$ are accessed from IMU. The policy outputs actuated joint position offsets as action vector $\boldsymbol{a}_t \in \mathbb{R}^{12}$. The target joint positions are determined by adding the joint offsets to the nominal joint angles $\boldsymbol{\theta}_0 \in \mathbb{R}^{12}$, which are transformed to joint torque by using joint PD controllers.

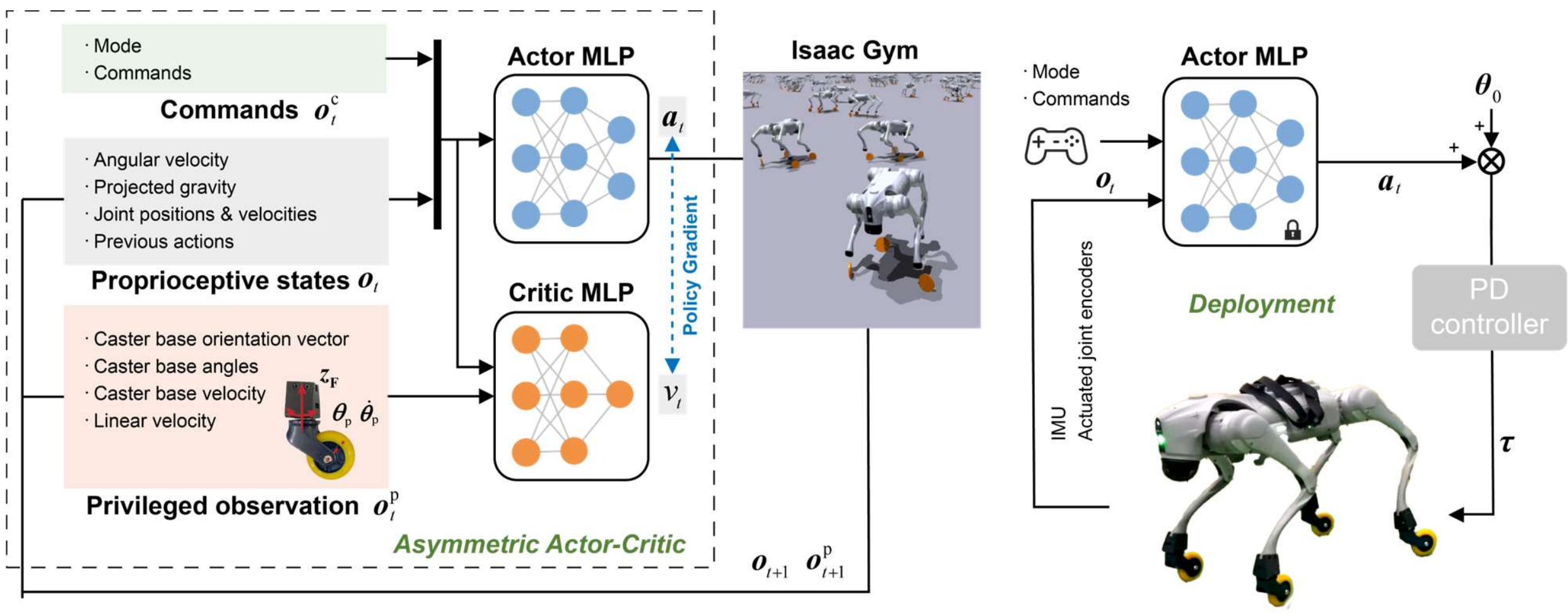

Fig. 4. The omnidirectional passive wheeled quadruped framework is based on an asymmetric Actor-Critic. During training, the privileged observations, including the wheel base axis posture vector $\boldsymbol{z}_{\mathrm{F}}$, angles $\boldsymbol{\theta}_{\mathrm{p}}$, and wheel angles $\dot{\boldsymbol{\theta}}_{\mathrm{p}}$ are obtained from the simulator in the training phase, which are not needed in deployment.

### B. Caster Base Posture Adjustment Strategy

Based on the propulsion mechanism analyzed above, the direction of the propulsive force generated by a passive caster is modulated by the posture of the caster base. Specifically, the constraint reaction induced by static friction can produce an effective propulsive component only when the swivel axis is properly inclined with respect to the ground. Therefore, the propulsive direction can be regulated by adjusting the caster base posture.

For this purpose, we employ the caster base posture adjustment strategy based on the reference velocity. Specifically, the angle $\boldsymbol{\theta}_{\mathrm{diffz}}$ between the horizontal component of the caster swivel axis vector $\boldsymbol{z}_{\mathbf{F}}^{xy}$ and the linear velocity vector $\boldsymbol{v}_{xy}^{\mathrm{ref}}$ expressed in the world coordinate system $\{I\}$ is constrained to satisfy $|\boldsymbol{\theta}_{\mathrm{diffz}}| < 90°$ to adjust the caster base's posture dynamically. This condition ensures that the horizontal projection of the swivel axis has a positive component along the desired direction of motion, thereby providing a directional bias that guides the learning policy toward generating propulsion in the commanded direction.

In addition, due to the eccentric distance between the rotation axis of the caster base and the wheel axle, a misalignment angle between the caster base moving direction and the wheel plane generates a torque around the base axis due to ground friction, compelling the caster base to rotate toward a coplanar configuration. We introduce angle deviation constraints $\boldsymbol{\theta}_{\mathrm{diff}\phi}$ between the caster base deflection angle $\phi$ and the command angle $\boldsymbol{\theta}_{{}^{\mathrm{b}}\boldsymbol{v}_{xy}^{\mathrm{ref}}} = \operatorname{atan2}({}^{\mathrm{b}}\boldsymbol{v}_y^{\mathrm{ref}}, {}^{\mathrm{b}}\boldsymbol{v}_x^{\mathrm{ref}})$ correspond to the reference velocity commands vector $\boldsymbol{v}_{xy}^{\mathrm{ref}}$ in the wheeled quadruped base coordinate system within 90 degrees.

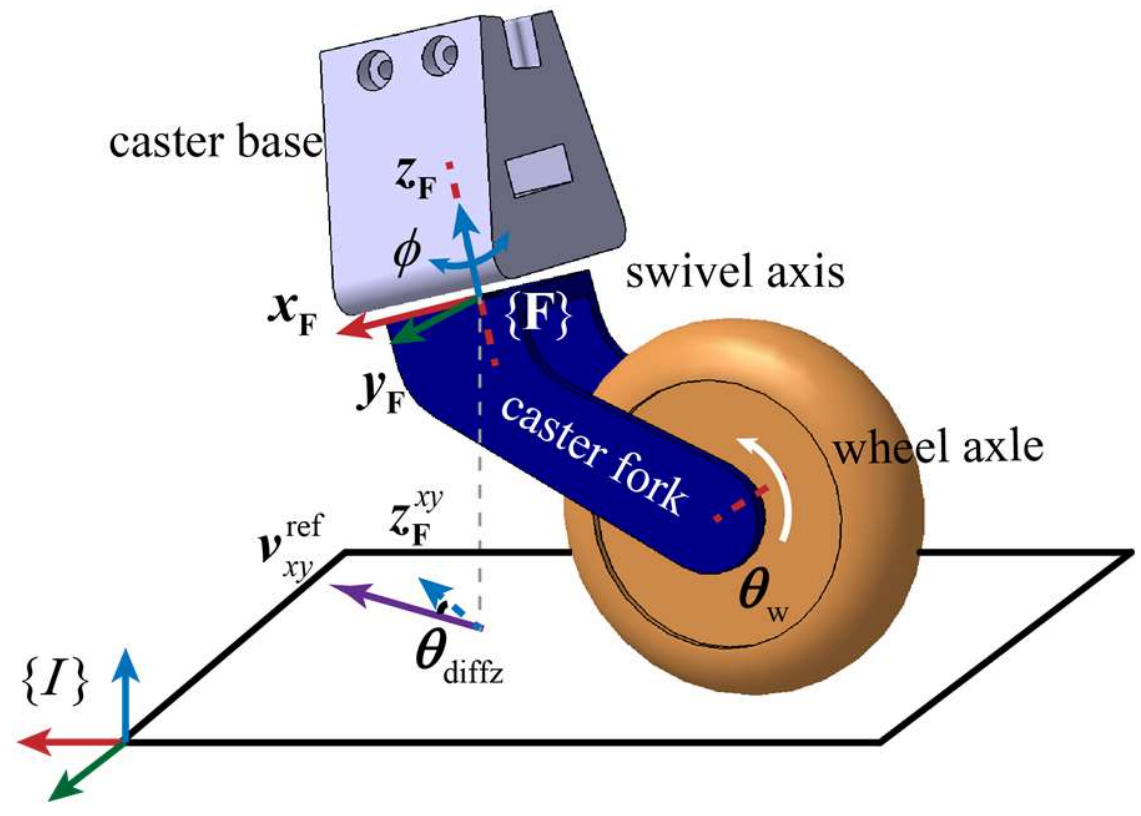

Fig. 5. The schematic diagram of caster base posture adjustment strategy. The coordinate base $\{\mathbf{F}\}$ is attached to the caster base.

### C. Observation Definition

The learning algorithm employed PPO under the Asymmetric Actor-Critic framework, providing different information to the Actor and Critic. The observation for the Actor, including velocity commands $\boldsymbol{o}_t^{\mathrm{c}} = (\boldsymbol{v}_x^{\mathrm{ref}}, \boldsymbol{v}_y^{\mathrm{ref}}, \boldsymbol{\omega}_z^{\mathrm{ref}})$ and proprioceptive states $\boldsymbol{o}_t$, denoted as $\boldsymbol{s}_t = (\boldsymbol{o}_t^{\mathrm{c}}, \boldsymbol{o}_t)$. The observation for the Critic, including privileged observation $\boldsymbol{o}_t^{\mathrm{p}}$ except for $\boldsymbol{s}_t$, denoted as $\tilde{\boldsymbol{s}}_t = (\boldsymbol{o}_t^{\mathrm{c}}, \boldsymbol{o}_t, \boldsymbol{o}_t^{\mathrm{p}})$. The specific expression is as follows:

$$\boldsymbol{o}_t = (\boldsymbol{\omega}_t, \boldsymbol{g}_t, \boldsymbol{\theta}_{\mathrm{a}_t}, \dot{\boldsymbol{\theta}}_{\mathrm{a}_t}, \boldsymbol{a}_{t-1})$$

$$o_t^{\mathrm{p}} = (z_{\mathrm{F}_t}, \phi_t, \dot{\phi}_t, \dot{p}_t)$$

where, $\omega_t \in \mathbb{R}^3$ and $g_t \in \mathbb{R}^3$ donate the angular velocity and gravity vector in the robot's base frame. $\theta_{\mathrm{a}_t} \in \mathbb{R}^{12}$ and $\dot{\theta}_{\mathrm{a}_t} \in \mathbb{R}^{12}$ represent the actuated joint positions and joint velocities, while $a_{t-1} \in \mathbb{R}^{12}$ is the previous action. $z_{F_t} \in \mathbb{R}^3$ indicate the $z$-axis of caster base coordinate frame $\{\mathbf{F}\}$ as shown in Fig. 5. $\phi_t$ and $\dot{\phi}_t$ represent the caster swivel axis joint positions and velocities, while $\dot{p}_t$ is the linear velocity observation of the torso in the world coordinate frame.

### D. Reward Function Design

The reward functions are designed to encourage passive wheeled quadruped robots to learn the posture adjustment strategy of the caster base, as shown in Section B, and to use the periodic oscillations motion of different numbers of redundant casters to achieve propulsion. We introduce a caster oscillations mask to encourage different casters to oscillate to achieve variable modes. The mask $\mathcal{M} = [\mathrm{b_{FL}}, \mathrm{b_{FR}}, \mathrm{b_{RL}}, \mathrm{b_{RR}}]$ is defined as an array of Boolean types. A value of $\mathrm{b}_i = 1$ represents a penalty for the oscillation of the *i*-th caster, while a value of $\mathrm{b}_i = 0$ indicates an encouragement for the *i*-th caster to oscillate. After obtaining the oscillating to propulsion ability, some reward functions are introduced to achieve velocity tracking tasks, including linear velocity $v_x^{\mathrm{ref}}$, $v_y^{\mathrm{ref}}$ tracking, and angular velocity $\omega_z^{\mathrm{ref}}$ tracking.

The caster base posture tracking reward, caster base joint axis angle tracking reward, and caster oscillation mask reward are only active when $\|v_{xy}\| > 0.2$. The set $D = \{\mathrm{FL, FR, RL, RR}\}$ indicates the four legs, and the set $P = \{\mathrm{wheel, caster\ base, calf}\}$ represents the links penalized for contact. The reward used in the training phase is shown in TABLE I.

TABLE I. REWARD FOR OMNIDIRECTIONAL PASSIVE WHEELED QUADRUPED

| *Reward Item* | *Definition* | *Weight* |
|---|---|---|
| Linear velocity tracking | $\exp\left(-\|v_{xy} - v_{xy}^{\mathrm{ref}}\|^2\right)$ | 2.0 |
| Angular velocity tracking | $\exp\left(-\|\omega_z - \omega_z^{\mathrm{ref}}\|^2\right)$ | 0.5 |
| **Caster base posture tracking** | $\max\left(\cos\left(\frac{\pi}{2}\right) - \cos(\theta_{\mathrm{diff}z}), 0\right)$ | 1.5 |
| **Caster base joint angle tracking** | $\min\left(\cos\left(\theta_{\mathrm{diff}\phi}\right) - \cos\left(\frac{\pi}{2}\right), 0\right)$ | -0.8 |
| **Caster oscillation mask** | $\sum_{i \in D} (\mathcal{M}_i \exp(-\|\dot{\phi}_i\|) + 0.5(1 - \mathcal{M}_i)(1 - \exp(-\|\dot{\phi}_i\|)))$ | 1.5 |
| Body height tracking | $\exp\left(-\|h - h^{\mathrm{ref}}\|^2\right)$ | 1.0 |
| Torso $z$-direction velocity | $-{v_z}^2$ | 2.0 |
| Torso roll-pitch velocity | $-\|\omega_{xy}\|^2$ | 0.05 |
| Collision penalty | $\sum_{i \in P} (\|f_i^{\mathrm{contact}}\| > 0.1)$ | -1.0 |
| Termination penalty | $\|f_{\mathrm{base}}^{\mathrm{contact}}\| > 0.1$ | -10.0 |
| Actuated joint acceleration penalty | $-\|\ddot{\theta}\|^2$ | -2.5e-7 |
| Actuated joint position limits | $-\|\theta_a - \mathrm{clip}(\theta_a, 0.9\theta_{\min}, 0.9\theta_{\max})\|$ | 1.0 |
| Actuated joint velocity limits | $-\max\left(\|\dot{\theta}_a\| - 0.9\dot{\theta}_{\mathrm{lim}}, 0\right)$ | 1.0 |
| Actuated joint torque limits | $-\max\left(\|\tau\| - 0.9\tau_{\mathrm{lim}}, 0\right)$ | 1.0 |
| Standing joint positions penalty | $-\|\theta_a - \theta^0\|$ | 0.8 |
| Action rate | $-\|a_t - a_{t-1}\|^2$ | 0.01 |

To bridge the sim-to-real gap, we employ domain randomization to enhance the adaptive performance of the control policy. Due to the influence of assembly precision on the casters, variations in joint damping significantly affect propulsion performance. To address this, we randomized the joint damping of the casters within the range [0.01, 0.1]. Additionally, the joint proportional gain (Kp) and derivative gain (Kd) were randomized within the range [0.9, 1.1]. To further improve the overall stability of the robot, the position of the torso center of mass was randomized within the range [-0.05, +0.05].

## V. EXPERIMENTS & RESULTS

### A. Caster Base Posture Adjustment Strategy Simulation

To validate the effectiveness of the caster base posture adjustment, a velocity-varying simulation experiment was designed, with the velocity command profile depicted in Fig. 6. The rotational angle of the passive caster base is directly obtainable from the simulation. Prior to 6 seconds, in the absence of velocity commands, the casters exhibited unordered motion to maintain chassis stability. From 6 to 10 seconds, as the robot was commanded to move backwards, $\cos(\phi)$ remained near $-1$, indicating that the caster base joint angle was approximately $\pm 180°$. During this period, both $\theta_{\mathrm{diff}z}$ and $\theta_{\mathrm{diff}\phi}$ maintained positive values, signifying that the posture $z_{\mathbf{F}}^{xy}$ of the caster base and its rotational angle $\phi$ formed an angle of less than 90° with the command vector. At 10 seconds, when the velocity command underwent a step change, the caster base angle transitioned from approximately $\pm 180°$ to near 0°. Concurrently, the error angles $\theta_{\mathrm{diff}z}$ and $\theta_{\mathrm{diff}\phi}$ briefly exceeded 90°, but the controller rapidly adjusted, reducing them to below 90° within approximately 0.5 seconds. These results demonstrate that the angle adjustment strategy for the caster base is highly effective and responsive. The entire velocity tracking process is illustrated in Fig. 7.

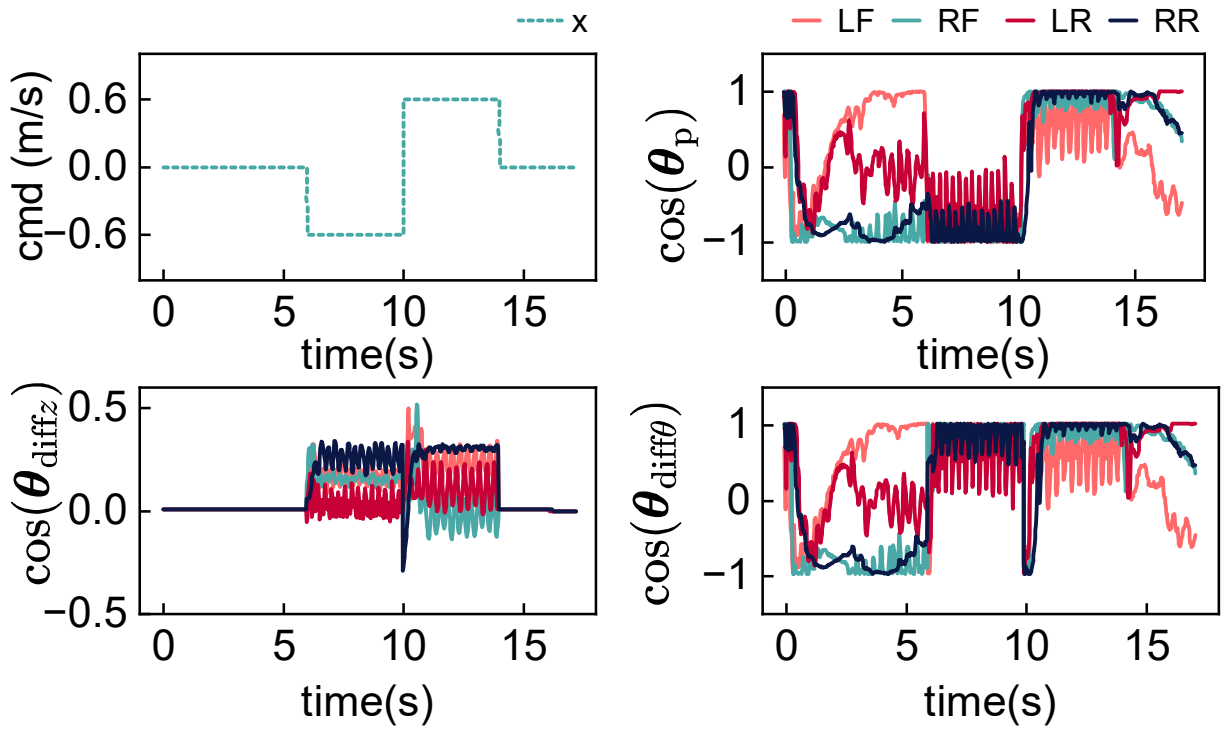

Fig. 6. The simulation results of the posture adjustment strategy.

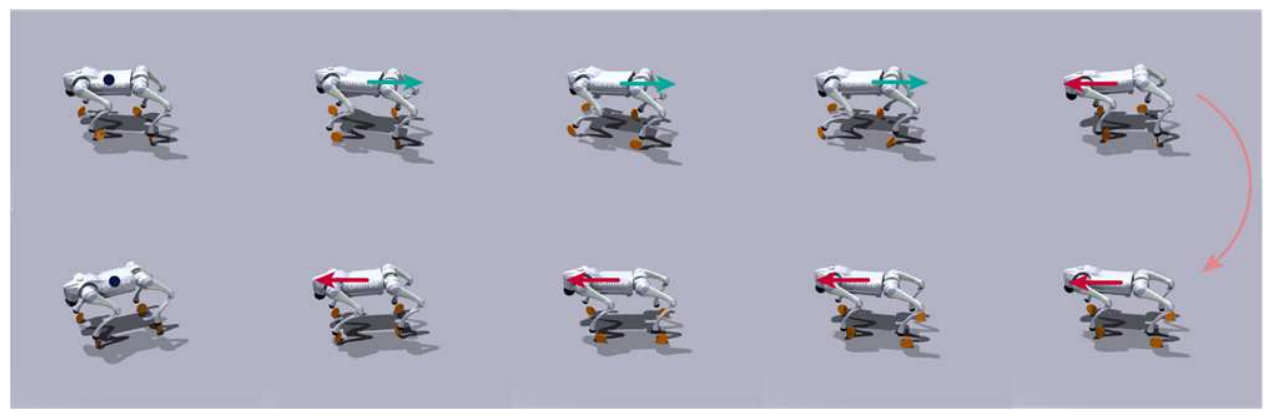

Fig. 7. The snapshots of the wheel base posture adjustment and wheel base angles relevant to the command. The blue arrows represent a command velocity of $-0.6\ \mathrm{m/s}$, and the red arrows represent a command velocity of $+0.6\ \mathrm{m/s}$.

### B. The Slalom Test in the Real World

To evaluate the omnidirectional mobility of the passive wheeled quadruped robot, a slalom test was designed, as illustrated in Fig. 1, with a spacing of 1.8 m between the poles. The result curves in Fig. 8 show that when the $y$-direction command is to the right, such as in the 4s to 7s interval, the lateral swing joints of the left legs (LF and FR) open to assist the robot's lateral motion. Conversely, when the y-direction command is to the left, the lateral swing joints of the right legs open. The omnidirectional passive wheeled quadruped robot's lateral swing joints, hip joints, and knee joints work in coordination to adjust the caster base posture and swing, enabling stable omnidirectional motion.

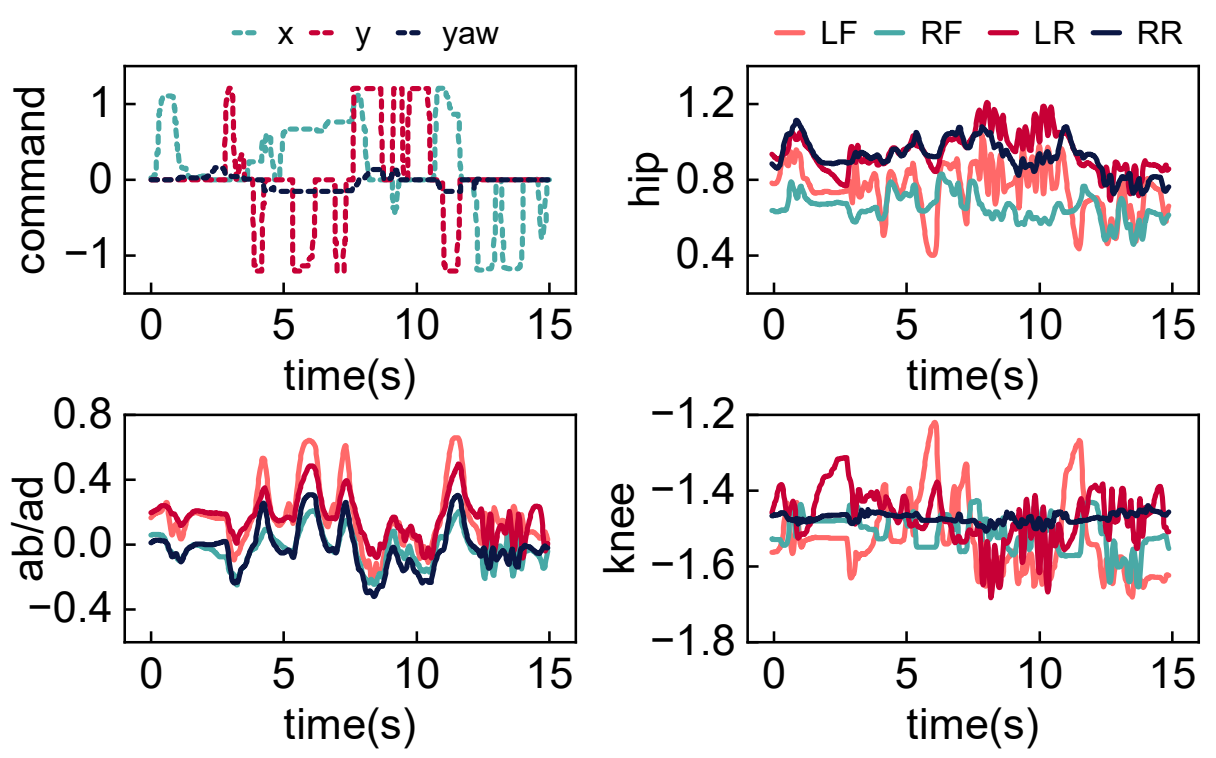

Fig. 8. The results of the actuated joint angles variation curves during the slalom test.

### C. High-speed Motion Experience

A high-speed movement test using a backwards motion was conducted on flat terrain, as shown in Fig. 9. The distance between two adjacent red floor tiles is 8.4 m, and the total time taken was 2.05s, yielding an average speed of approximately 4.1 m/s. The robot exhibited low energy consumption during the constant-velocity phase due to incorporating passive casters.

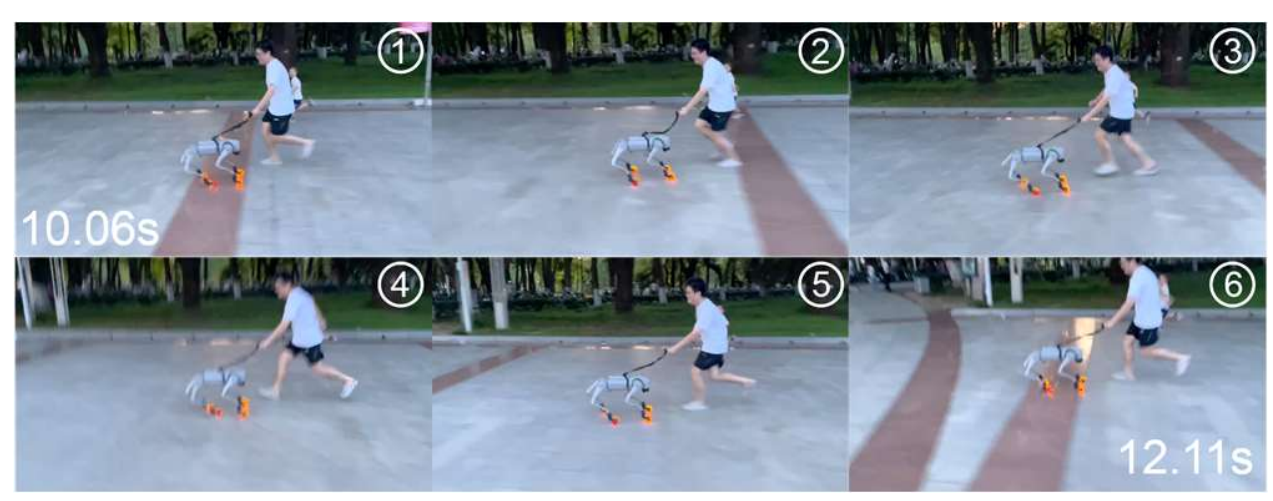

Fig. 9. The Snapshots of high-speed motion.

### D. Mode Switch of Passive Wheeled Quadruped Robots

Since the casters mounted on the four legs are redundant for propulsion, we designed various drive modes. In practical experiments, we validated the propulsion and switching of three modes: four-wheel drive, front-wheel drive, and rear-wheel drive, as depicted in Fig. 10. After applying a mask to disable specific propulsion legs, the linear velocity at the foot end was correspondingly reduced.

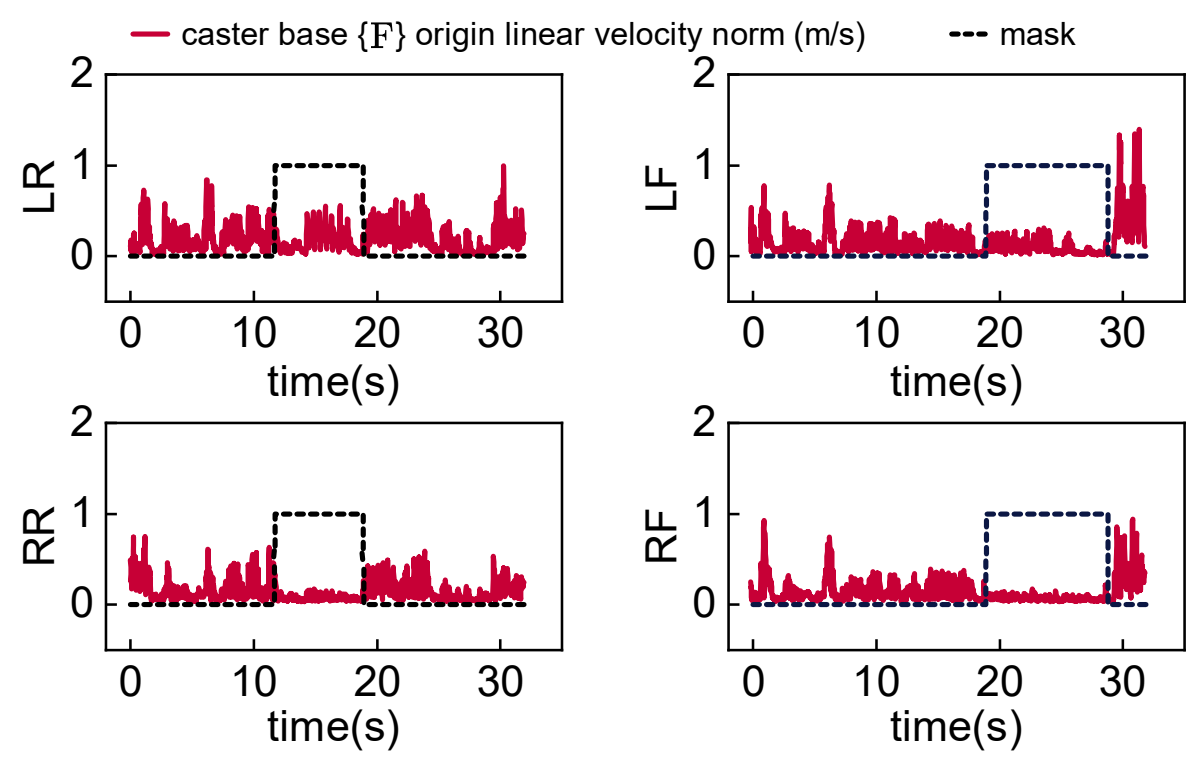

Fig. 10. The variable motion mode switch results from masking the legs' twisting oscillatory.

### E. Cost of Transport Comparison

To compare the energy consumption performance of different locomotion approaches, the cost of transport (COT) is taken into account as an evaluation indicator. The mechanical COT is the mechanical positive work performed per unit mass per unit distance, which is defined by

$$COT_{\text{mech}} = \frac{h_k \sum_{k=2}^{N} \sum_{i}^{4} \max\ (\tau_{i,k}\dot{\theta}_{i,k}, 0)}{m_0 \mathrm{g}(x_{N-1} - x_0)}$$

where $h_k$ represents the time interval between the $k$-th node and the $(k+1)$-th node, $N$ denotes the total number of nodes, $\tau_{i,k}$ indicates the driving torque of the $i$-th actuated joint at the $k$-th node, and $\dot{\theta}_{i,k}$ represents the joint velocity of the $i$-th actuated joint at the $k$-th node. $x_0$ and $x_{N-1}$ respectively denote the displacement of the robot in the forward direction at the initial time and the termination time.

We compared the COT of the same quadruped robot operating under two distinct locomotion modes, point-foot and caster, while traversing an identical trajectory on campus. The experiment trajectory length was 1.1 kilometers, as depicted in Fig. 11.

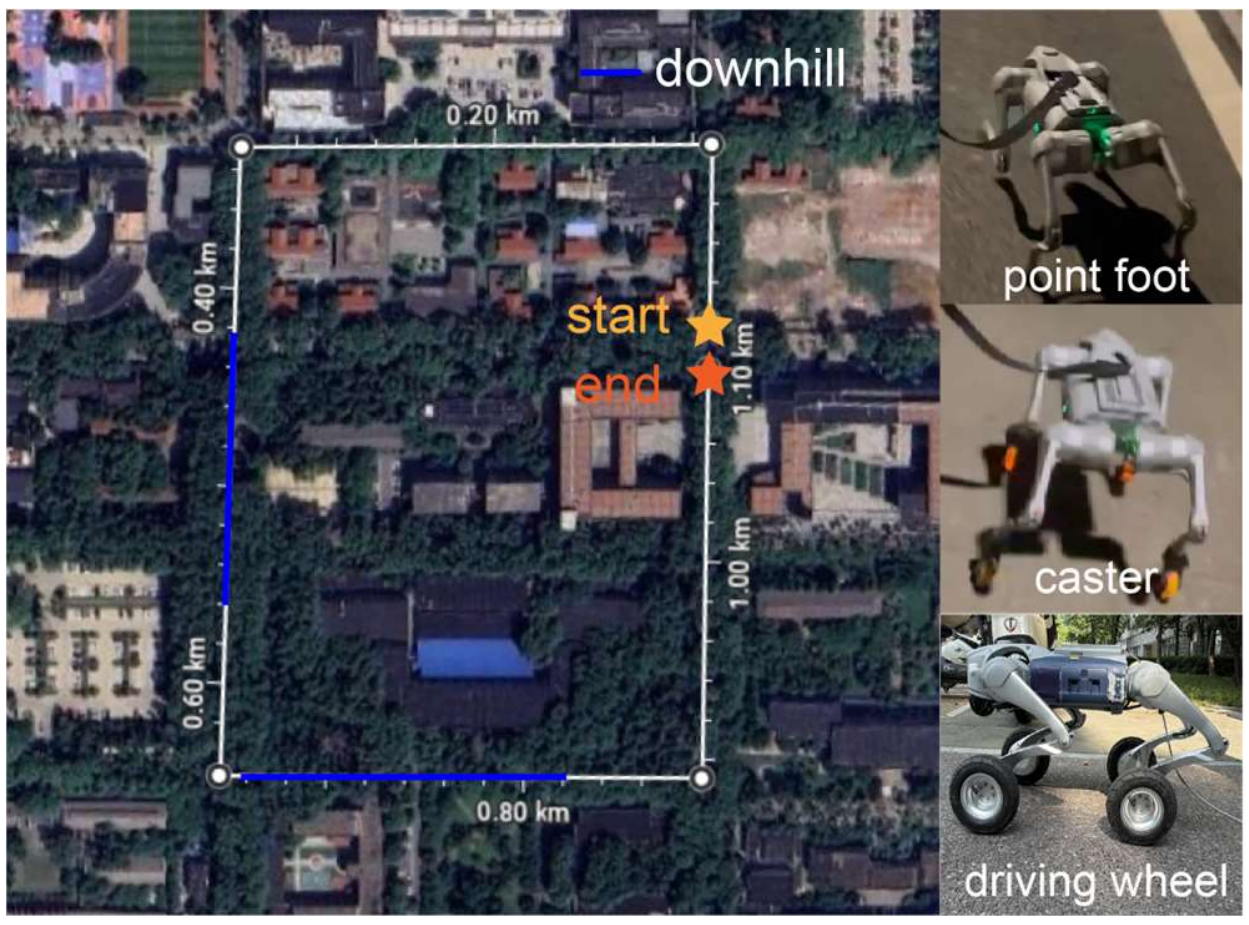

Fig. 11. The experiment trajectories of a quadruped on campus. The blue curve is downhill terrain.

TABLE II. Comparison of Point Foot Trot Locomotion and Passive Caster Locomotion

| *Propulsion Type* | *Avg. Battery Output Power* (W) | *Avg. Velocity* (m/s) | *Mechanical COT* |
|---|---|---|---|
| Point foot | 334.93 | 1.124 | 0.648 |
| Caster foot | **192.69** | **1.414** | 0.0705 |
| Driving wheel | 229.26 | 0.842 | **0.0637** |

TABLE II summarizes the results with two locomotion methods on the same quadruped Go2 from Unitree Robots. The test trajectory includes two segments of downhill terrain, as shown in Fig. 11 The remaining sections consist of flat ground and uphill terrain. As can be seen in TABLE II, the passive wheeled quadruped robot decreases the COT by approximately 89.1% w.r.t. the point foot robot's trotting mode. In terms of energy consumption, the passive wheeled-legged robot consumed approximately 41Wh to complete the experiment trajectory, significantly lower than the 91Wh consumed by the point-foot robot.

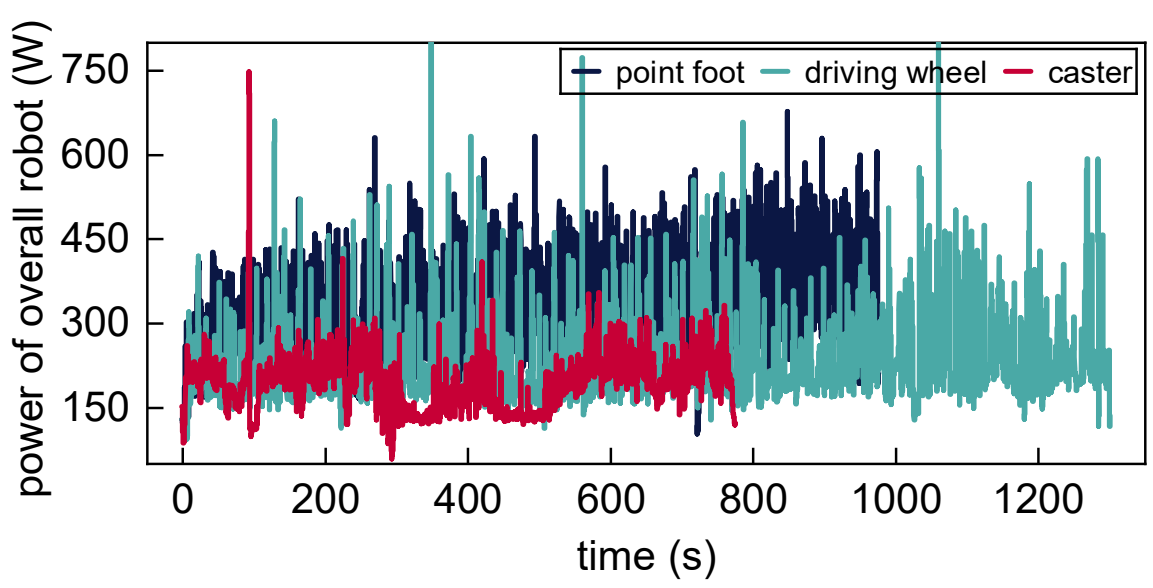

Fig. 12. The power curves of the caster and the point foot in the same experiment trajectory. The yellow background indicates that the test trajectory of the passive wheeled quadruped robot is on a downhill section.

As shown in Fig. 12, the power consumption of the passive wheeled quadruped robot decreases significantly on downhill terrain, whereas point-foot robots lack this advantage. Even on uphill segments, the energy consumption of the proposed passive wheeled-legged robot remains substantially lower than that of point-foot locomotion.

### *F. The*

## VI. Conclusion

This paper presents a novel omnidirectional passive wheeled-legged robot that generates propulsion through friction between the ground and 2-DOF casters mounted at each leg end. We propose a caster base posture adjustment strategy that leverages each leg's actuated DOF to adjust the caster base posture, thereby modifying the propulsion direction to achieve omnidirectional mobility. Additionally, we developed multiple drive modes for the robot, enabling propulsion through varying numbers of caster oscillations while ensuring stable transitions between modes. Our approach enables the omnidirectional passive wheeled quadruped to perform a slalom test, achieve high-speed motion up to 4 m/s, and reduce the COT by more than 89% compared to traditional walking gaits. A current limitation of the system is the suboptimal velocity tracking performance due to the absence of caster angle sensors. We plan to integrate visual odometry into the control loop to achieve precise velocity tracking in future work.

## Acknowledgment

This work was partially supported by the National Natural Science Foundation of China (No. 52375014), Guangdong Innovative and Entrepreneurial Research Team Program (No. 2019ZT08Z780), Dongguan Introduction Program of Leading Innovative and Entrepreneurial Talents (No. 20181220), and Wuhan Science and Technology Major Project (No. 2024060902020442).